\documentclass[conference]{IEEEtran}
\IEEEoverridecommandlockouts
\usepackage{cite}
\usepackage{amsmath,amssymb,amsfonts}
\usepackage{algorithmic}
\usepackage{graphicx}
\usepackage{textcomp}
\usepackage{xcolor}
\def\BibTeX{{\rm B\kern-.05em{\sc i\kern-.025em b}\kern-.08em
    T\kern-.1667em\lower.7ex\hbox{E}\kern-.125emX}}

\makeatletter
\newcommand{\linebreakand}{
  \end{@IEEEauthorhalign}
  \hfill\mbox{}\par
  \mbox{}\hfill\begin{@IEEEauthorhalign}
}
\makeatother
    
\begin{document}

\title{RaceLens: A Machine Intelligence-Based Application for Racing Photo Analysis\\
}

\author{
\IEEEauthorblockN{1\textsuperscript{st} Andrei Boiarov}
\IEEEauthorblockA{
\textit{Constructor Technology}\\
Schaffhausen, Switzerland \\
andrei.boiarov@constructor.tech}
\and
\IEEEauthorblockN{2\textsuperscript{nd} Dmitry Bleklov}
\IEEEauthorblockA{
\textit{Constructor Technology}\\
Schaffhausen, Switzerland \\
dmitry.bleklov@constructor.tech}
\and
\IEEEauthorblockN{3\textsuperscript{rd} Pavlo Bredikhin}
\IEEEauthorblockA{
\textit{Constructor Technology}\\
Schaffhausen, Switzerland \\
c-pavlo.bredikhin@constructor.tech}
\and
\IEEEauthorblockN{4\textsuperscript{th} Nikita Koritsky}
\IEEEauthorblockA{
\textit{Constructor Technology}\\
Schaffhausen, Switzerland \\
nikita.koritsky@constructor.tech}
\and
\IEEEauthorblockN{5\textsuperscript{th} Sergey Ulasen}
\IEEEauthorblockA{
\textit{Constructor Technology}\\
Schaffhausen, Switzerland \\
su@constructor.tech}
}

\maketitle

\begin{abstract}
This paper presents RaceLens, a novel application utilizing advanced deep learning and computer vision models for comprehensive analysis of racing photos. The developed models have demonstrated their efficiency in a wide array of tasks, including detecting racing cars, recognizing car numbers, detecting and quantifying car details, and recognizing car orientations. We discuss the process of collecting a robust dataset necessary for training our models, and describe an approach we have designed to augment and improve this dataset continually. Our method leverages a feedback loop for continuous model improvement, thus enhancing the performance and accuracy of RaceLens over time. A significant part of our study is dedicated to illustrating the practical application of RaceLens, focusing on its successful deployment by NASCAR teams over four seasons. We provide a comprehensive evaluation of our system's performance and its direct impact on the team's strategic decisions and performance metrics. The results underscore the transformative potential of machine intelligence in the competitive and dynamic world of car racing, setting a precedent for future applications.
\end{abstract}


\begin{figure*}[t]
  \centering
  \includegraphics[width=0.9\textwidth]{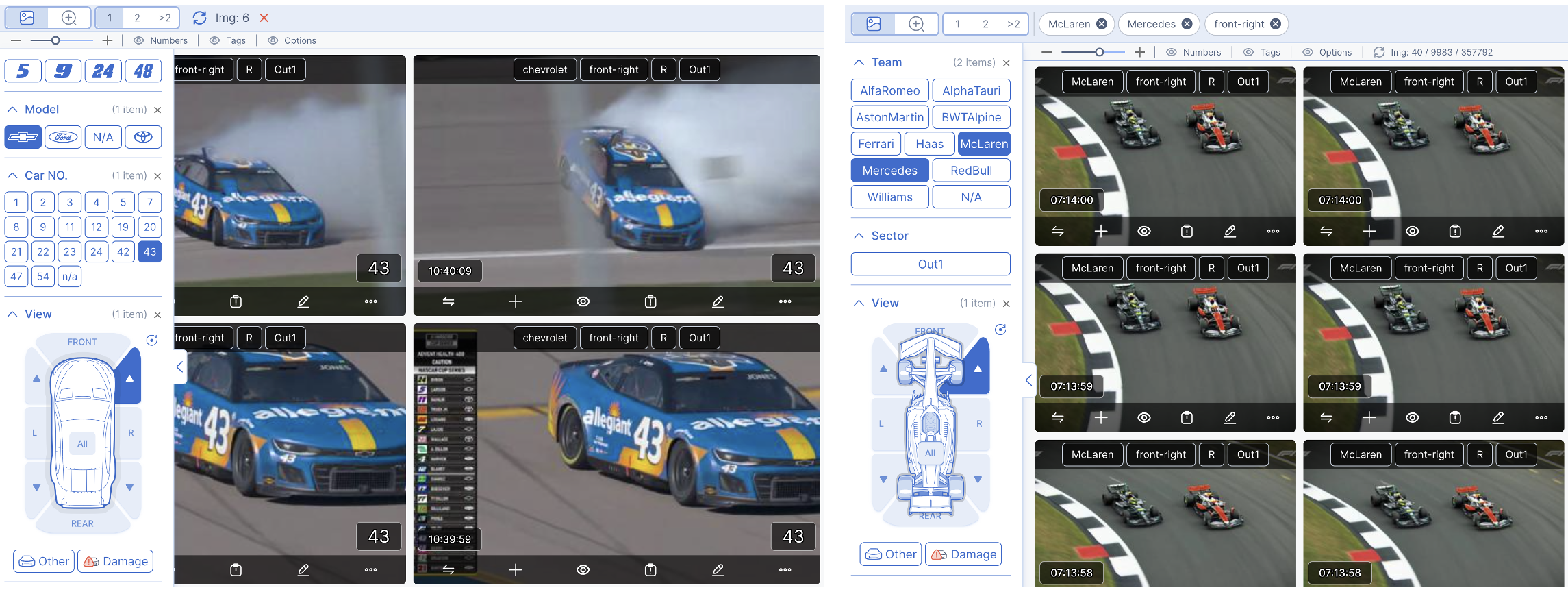}
  \caption{RaceLens interface examples. Left: NASCAR car with number 43, Chevrolet manufacturer and front-right orientation. Right: Formula 1 cars from two teams with front-right orientation.}
  \label{fig:interface}
\end{figure*}

\section{Introduction}

The utilizing of Machine Learning (ML) and Computer Vision (CV) approaches has marked a new era in various fields of real-world application, revolutionizing the way we process and interpret data. One such domain that is starting to reap the benefits of these advanced technologies is motorsports, where enormous volumes of data are generated and where rapid and accurate interpretation can significantly influence race teams performance and outcomes.

In the highly competitive world of motorsports, like NASCAR, even the slightest edge in strategy and decision-making can result in substantial advantages. For years, teams have been harnessing data from a variety of sources, including telemetry systems and trackside observations. However, one data source has been relatively untapped until now: racing photos. These snapshots contain a wealth of information about the race, the competitors, and the conditions, but until recently, extracting this data was a manual, time-consuming, and often error-prone process.

This paper presents RaceLens, an application that leverages deep learning and computer vision models to automatically analyze racing photos. Its user interface is shown in the Fig.~\ref{fig:interface}. Built with an aim to maximize the potential of racing photographs, RaceLens demonstrates proficiency in identifying and interpreting crucial elements in the images, such as detecting racing cars, recognizing car numbers, detecting and quantifying car details, and recognizing car orientations. Our work covers the development and implementation of RaceLens, from collecting and refining a comprehensive dataset for model training to deploying the application in a real-world scenario. We outline how a feedback loop between the application output and dataset augmentation significantly enhances the overall performance and accuracy of RaceLens over time. Perhaps most significantly, we provide an account of the successful deployment of RaceLens by several NASCAR teams over four seasons. By analyzing the impact of RaceLens on the team's performance, we illustrate the transformative potential of ML and CV in the high-stakes world of professional car racing. As illustrated in the Fig.~\ref{fig:interface}, RaceLens can successfully work with photos from different racing series (NASCAR, Formula 1, etc.).


The RaceLens application consists of 4 main modules: race car components detection, race car orientation recognition, race teams identification, race car detail automatic measurement. This paper is organized as follows: Section 2 provides a comprehensive description of the application modules; Section 3 presents the results of our case study with NASCAR teams; and Section 4 concludes the findings and discusses future directions for this research.

\section{RaceLens modules}\label{sec:modules}

\subsection{Race car components detection.}\label{sec:detect}

\textbf{Application feature.} The detection of a race car and its main attributes (like number and manufacturer's brand) are essential tasks in the whole RaceLens application. All subsequent application features described in the paper are built on the cropped image of the detected car. Detection of a car number is necessary to identify the car in the race, and car manufacturer's brand detection is important to reduce the error of car identification since different teams are using cars from different manufacturer's brands.

\textbf{ML approach.} To solve these detection tasks we are using deep learning based object detector model EfficientDet~\cite{b1} trained on different datasets. One model was trained for racing cars detection, another --- for racing cars attributes (car number and manufacturer's brand) detection. EfficientDet architecture was selected due to it inference speed and accuracy. The models achieve these properties by using bi-directional feature network (BiFPN) and special scaling rules: each network component, i.e., backbone, feature, and box/class prediction network, have a single compound scaling factor that controls all scaling dimensions using heuristic-based rules. The model inference speed is crucial for RaceLens because it is necessary to process photos in the nearly real-time regime.

Since EfficientDet is an anchor-based method, to train it for each task we used k-means clustering algorithm on each dataset to find aspect ratios of anchor boxes. During training 3-channel images were resized to 512x512, AdamW optimizer~\cite{loshchilov2018decoupled} with learning rate $0.001$ and cosine annealing learning rate schedule were used.

To recognize detected car numbers we developed a separate model. This model takes a cropped number from the results of attributes detection model and predicts a car number. The main problem in training such kind of model is that not all car numbers are available in the train set as well as fonts and colors of these numbers may differ from race to race. To handle these challenges we proposed the following approach: we used heuristic algorithm to find image patches with separate digits, then each of these patches is processing with EfficientNet~\cite{tan2019efficientnet} to predict the digit. To train this digit recognition model the cross-entropy loss function is used. On inference all predicted digits a combined to produce final car number prediction.

To recognize manufacturer's brands was also trained a separate model. This model is a lightweight EfficientNet backbone with classification head for 4 classes.

\textbf{Dataset description.} We collected the proprietary dataset with NASCAR racing photos. Then for each task we constructed special datasets by selecting appropriate images and labeling with professional annotators. Final datasets descriptions are in Table~\ref{tab:car_detection_dataset}. We are constantly increasing our datasets based on new race events and users feedback. It allows us to avoid data drift effect. User can add image into feedback from the RaceLens application interface. Corner case examples are adding into the test set that are using for offline metrics calculation.

Numbers recognition dataset includes many heavy tailed classes. To improve classes balance we are using synthetic data generation. We trained the BigGAN~\cite{brock2018large} to generate about $0.2\%$ of the dataset images which increased offline metric (mAP) of the corresponding model by $0.5\%$.

\begin{table}
    \centering
    \caption{Race car components detection datasets}
    \begin{tabular}{|c|c|c|c|}
        \hline
        \textbf{Task} & \textbf{Classes} & \textbf{Train images} & \textbf{Test images} \\
        \hline
        Car detection & 1 & 16 524 & 1537 \\
        \hline
        Car attributes detection & 2 & 14 581 & 1570 \\
        \hline
        Numbers recognition & 72 & 843 100 & 40 114 \\
        \hline
        Manufacturers recognition & 4 & 222 908 & 1182 \\
        \hline
        \end{tabular}
    \label{tab:car_detection_dataset}
\end{table}

\textbf{Offline metrics.} As was discussed in the previous section, we constructed test sets for each of our models. After models training we measure offline metrics on these tests and make a decision of models updating on the production side. For detection tasks standard mean average precision (mAP) is used, for recognition --- accuracy metric. Results of all race car components models can be found in Table~\ref{tab:car_detection_metrics}. 

\begin{table}
    \centering
    \caption{Race car components models offline metrics}
    \begin{tabular}{|c|c|}
        \hline
        \textbf{Model} & \textbf{mAP / Accuracy} \\
        \hline
        Car detection & 95.2 \% \\
        \hline
        Car attributes detection & 91.3 \% \\
        \hline
        Numbers recognition & 98.6 \% \\
        \hline
        Manufacturers recognition & 99.7 \% \\
        \hline
        \end{tabular}
    \label{tab:car_detection_metrics}
\end{table}

\subsection{Race cars orientation recognition}\label{sec:car_orientation}

\textbf{Application feature.} The car orientation model developed for RaceLens aims to accurately estimate the orientation of cars in NASCAR Cup races. By analyzing photo streams from different cameras placed around the race track, the model provides the opportunity to cluster different cars, reducing the time for teams to match cars during comparison.

\textbf{ML approach.} The car orientation model is a multi-class classification model that predicts an array of probabilities for 8 classes: front, front-left, front-right, rear, rear-left, rear-right, left and right. It is based on the EfficientNet model with a few modifications. The car orientation model uses the results of the race car detection model described in Section~\ref{sec:detect}. It takes a cropped 3-channel image of the car with a shape 100x200 as input and returns an array with probabilities for each orientation. During training, the model utilized the AdamW optimizer and a scheduler that reduces the learning rate when a metric has stopped improving. The initial learning rate was set to $0.01$, with a reduction factor of $0.1$. Class balancing techniques were applied to handle imbalanced data.


\textbf{Dataset description.} To train, evaluate and test the car orientation model, a comprehensive dataset of NASCAR photos was collected. The dataset includes over 100 000 images from more than 100 racing events spanning from 2018 to 2022 labeled by professional annotators. The images consist of cars with different paint schemes and were captured under various conditions such as night, rain, and sunny conditions. The dataset was divided into $70\%$ for training, $10\%$ for validation, and $20\%$ for testing.

\textbf{Offline metrics.} Offline tests were conducted to evaluate the performance of the car orientation model. The model achieved high accuracy for each orientation class, as shown in Table~\ref{tab:test_car_orientation}. These results demonstrate the effectiveness of the car orientation model in accurately predicting the orientation of cars in NASCAR races.

\begin{table}
    \centering
    \caption{Race cars orientation recognition offline tests}
    \begin{tabular}{|c|c|}
        \hline
        \textbf{Class} & \textbf{Accuracy } \\
        \hline
        Front & 97.3 \% \\
        \hline
        Front-left & 97.6 \% \\
        \hline
        Front-right & 99.4 \% \\
        \hline
        Left & 98.0 \% \\
        \hline
        Rear & 98.0 \% \\
        \hline
        Rear-left & 99.2 \% \\
        \hline
        Rear-right & 99.8 \% \\
        \hline
        Rear & 97.3 \% \\
        \hline
    \end{tabular}
    \label{tab:test_car_orientation}
\end{table}

\subsection{Race cars teams identification}

\textbf{Application feature.} Filtering car images by team is a crucial aspect of the RaseLens user workflow. 
This task relies on visual pattern recognition since each team has a unique color scheme for their car. 
By leveraging advanced algorithms and machine learning techniques, users can efficiently categorize and organize car images based on team affiliation.  This functionality streamlines workflow, enhances data organization, and provides valuable insights into team performance and strategies within the RaseLens platform.

\textbf{ML approach.} A Metric Learning approach is employed to tackle the task.
The main encoder model takes a 3-channel image as input and outputs a 1-D vector (embedding) that represents the color scheme of the car in the image. 
The embeddings are trained to be closer to each other for images of the same class (team) and farther apart for different classes.  The cosine distance metric is used to measure the closeness of embeddings. During the training phase, a triplet loss~\cite{hoffer2015deep} is utilized to minimize the distance between embeddings of similar class images and maximize the distance between embeddings of different class images. 
Additionally, a fully connected layer with cross-entropy loss is employed to improve the results.

In the inference phase, clusters can be created using the embeddings, but there is no information about the corresponding team names. To address this, the Car Number Recognition Model (CNRM) described in Section~\ref{sec:detect} is utilized. During the race, images are processed one by one. If the CNRM provides high accuracy in number detection, the image embedding is extracted and used as a reference embedding. Once the number of reference images for a given class reaches a certain threshold, a centroid embedding is calculated by averaging the reference embeddings. From this point forward, if the distance from an image embedding to a class centroid is below a threshold, the class is assigned to the image. This approach allows for clustering of images based on color scheme and uses the CNRM to assign the corresponding team names to the clusters during the inference phase.

\textbf{Dataset description.} The primary data source for the project is professional race photographs taken during the event. From each frame of these photographs, all cars are detected, and the corresponding crops are saved. These crops are then labeled semi-automatically with numbers using a CNRM and internal labeling validation tools. Through several iterations of labeling and validation, a high-quality dataset is obtained, ensuring accurate and reliable data for further analysis and model training.

\textbf{Offline metrics.} For this task we use the following custom metrics. \textit{Mean cluster deviation (MClD)}: In each cluster distance from all embeddings to centroid is calculated, then averaged. Metric depicts mean of this averages. Lower metric value is better. \textit{Mean centroid deviation (MCeD)}: Average distance among all the centroids. Higher metric value is better. \textit{Mean intra-outra distances delta (MIODD)}: This metric does not use centroid for calculation. First, for each class distances between embeddings of given class is calculated and averaged (intra-). Second, for each pair of classes distances between different classes are calculated and averaged (outra-). Finally, delta between this intra- and outra- distances depicts the separating ability of the model.  

Our model provides the following results: MClD = 0.167, MCeD = 1.813, MIODD = 0.572 on the test dataset of 14 867 images.

\subsection{Race cars details automatic measurement}

\textbf{Application feature.} Accurate measurement of car details sizes for RaceLens users requires establishing a direct correlation between image distances (in pixels) and real-world distances (in millimeters / inches). 
This is achieved by detecting specific persistent parts of the car with precision. 
Using a standardized wheel disk size is convenient for NASCAR racing cars as it allows for consistent measurements across multiple images.

\textbf{ML approach.} The described pipeline utilizes multiple models. 
Auxiliary models are responsible for detecting wheel bounding boxes and car orientation (described in~\ref{sec:car_orientation}). They operate on all frames and do not require pixel-wise accuracy.
The wheel keypoints model is specifically designed to work on cars in side position, preferably in the pitstop line. This positioning ensures that the wheels are mostly in profile, minimizing perspective distortions.
The wheel crops obtained from the previous step are then fed into the Keypoint R-CNN with ResNet-50 backbone~\cite{he2016deep}. This model is based on Mask R-CNN~\cite{he2017mask} architecture which is complemented by a head that predicts coordinates of 6 keypoints: 4 edges of the wheel disk (top, right, bottom, left), the disk center, and the point where the wheel touches the ground. The loss function used for this task is the mean-squared error, the input of the model is a 3-channel 512x512 image. With these keypoints detections, the radius of the wheel disk is calculated, and two lines are drawn: one connecting the centers of the disks and another connecting the points where the wheels touch the ground. Since the size of the wheel disk radius is known in advance, distances such as line lengths can be automatically calculated based on this information.

\textbf{Dataset description.} The dataset collection process for auxiliary models is similar to other models. 
The keypoints dataset includes car wheels crops with six keypoints coordinates, and if the point of ground touch is not available on the wheel crop, the coordinate $(image\_width / 2, 0)$ is used instead. The train set includes 2868 images.

\textbf{Offline metrics.} The evaluation for the dataset utilizes COCO metrics~\cite{lin2014microsoft}, which are based on the concepts of bounding box evaluation. The main metrics are Average Precision (AP) and Average Recall (AR), with values of 0.977 and 0.986 respectively on the test dataset of 288 images. 

\section{Production results}

The RaceLens is used by NASCAR teams during 2018. Our application is deployed for team on Azure virtual machine with 4 vCPU with 16GB RAM and 1 vGPU Nvidia Tesla K80 with 12GB memory, data is storing in MongoDB, deep learning models described in Section~\ref{sec:modules} are trained and deployed with PyTorch library~\cite{pytorch}. During the operating period more than 200 race events were processed with RaceLens and this amount keeps growing, the average number of photos processed during the race event is 7000. 

The main online metrics that characterize work quality of our application are percent of photos without cars (N/A photos) and percent of photos labeled by users as a feedback (Feedback photos). Since the photos during a race event are taken by professional photographers in order to get images of certain cars in a certain orientation, the number of photos in which there are no cars is extremely small. The average percent of N/A photos in race that RaceLens returns is less than $1 \%$. This number is perfectly match with a true distribution of such photos during a race. The average percent of Feedback photos in race event is $1 \%$. This number does not affect the user experience, and we use these feedback images to update the test sets.    


\section{Conclusion}

In this paper, we presented RaceLens, a machine intelligence-based application designed for racing photo analysis, leveraging state-of-the-art deep learning and computer vision techniques. Our experimental results demonstrated the ability of our models to perform a wide range of tasks such as detecting cars, recognizing car numbers, detecting car attributes, and assessing car orientations. Our approach for augmenting and improving the dataset, combined with a feedback loop for continuous model improvement, allowed RaceLens to continuously evolve, enhancing its performance and accuracy over time. The successful deployment of RaceLens at NASCAR teams over four seasons, provided clear evidence of the application's practical utility.

Despite its successes, we recognize that there is room for improvement and expansion. Future work could explore the use of RaceLens in other forms of racing and integrate it with other types of real-time data feeds to provide instant analysis and strategic guidance.

\end{document}